# An LLM Maturity Model for Reliable and Transparent Text-to-Query

**Lei Yu and Abir Ray**
Expression
1140 3rd St NE #310, Washington, DC 20002
lyu@expr.ai, aray@expr.ai

## Abstract

Recognizing the imperative to address the reliability and transparency issues of Large Language Models (LLM), this work proposes an LLM maturity model tailored for text-to-query applications. This maturity model seeks to fill the existing void in evaluating LLMs in such applications by incorporating dimensions beyond mere correctness or accuracy. Moreover, this work introduces a real-world use case from the law enforcement domain and showcases QueryIQ, an LLM-powered, domain-specific text-to-query assistant to expedite user workflows and reveal hidden relationship in data.

## 1 Introduction

Large Language Models (LLM) [7] have evolved into powerful tools with wide-ranging applications, and among these, text-to-query has emerged as a critical and rapidly advancing direction. The ability of LLMs to convert user questions and instructions in natural languages to actionable queries in structured query languages (SQL) or NoSQL holds immense potential for streamlining information retrieval processes and unleashing data at speed and scale. As the impact of LLMs on business operations and our daily lives continues to grow, the importance of ensuring their reliability and transparency becomes paramount. Existing studies on text-to-SQL [3,5,6] and benchmarks such as the Spider leaderboard [2], predominantly focus on accuracy metrics. These evaluations often overlook the broader spectrum of the trustworthiness of LLMs [1,4], leaving gaps in our understanding of LLM performance and maturity.

This work proposes an LLM maturity model which is distinguished by its focus on three pivotal questions that delve into the reliability and transparency of LLMs in text-to-query applications. The first question addresses the fundamental concern of reliability: *Does the LLM produces correct results*? This inquiry encompasses considerations of accuracy, factuality, and the prevention of hallucinations in LLMs. The second question probes into the consistent and sustained performance of LLMs: *Does the LLM consistently produce correct results*? While correctness is paramount, our approach extends beyond it to scrutinize the nuanced facets of reliability, ensuring that LLMs not only generate correct outputs but do so consistently and without introducing confusion or misinformation. The third question centers on transparency: *Does the LLM explain and justify the results it claims to be correct*? This question delves into concerns around interpretability, explainability, observability, and traceability, providing insights into how well LLMs articulate and present their decision-making processes to users and stakeholders.

In this paper, we deliberately narrow our focus to these three key questions, emphasizing their significance in addressing reliability and transparency concerns in the broad landscape of trustworthy AI. We acknowledge that other important dimensions, such as privacy, security, safety, ethics, fairness, and costs, merit in-depth studies for the development of LLM maturity models, but they fall beyond the scope of this work.

Section 2 introduces a real-world use case from the law enforcement domain and illustrates the text-to-query capability of a fine-tuned LLM in QueryIQ, an LLM-powered, domain-specific text-to-query assistant developed by Expression. Section 3 presents the proposed LLM maturity model with an overview and detailed breakdown of characteristics and acceptance criteria across increasing levels of maturity. Section 4 highlights some future directions along LLM maturity models.

## 2 Text-To-Query Use Case

The text-to-query capability enabled by LLMs has the potential to unleash data at speed and scale by democratizing data access and analytics for all. We show a use case from the law enforcement domain where we employ an LLM-powered, domain-specific QueryIQ assistant to expedite user workflows and reveal non-obvious relationship in data. This illustrative use case has corollary to a wide range of text-to-query use cases and workflows across a variety of domains and applications.

Imagine a scenario where a law enforcement agent obtains a phone number from a source and seeks to: 1) determine if the number is linked to any suspects in a database of individuals with criminal records; 2) upon finding a match, identify the top N individuals who have had the most phone calls with the suspect; 3) uncover any notable patterns among these individuals when plotted on a map based on their addresses. Empowered by our QueryIQ assistant, the agent can effortlessly inquire about complex data without prior knowledge of the underlying data sources or how to formulate database queries. Through a simple chat interface, the agent can pose a series of questions and immediately receive precise results in desired formats, with hidden relationships visualized on a map. Additionally, QueryIQ ensures traceability and transparency, allowing the gent to trace intermediate steps leading to answers and examine the database queries generated by the underlying LLM to retrieve relevant data elements. This enables agents to iteratively prompt the system with further inquiries, facilitating deeper investigation. Below are examples of queries an agent could pose to QueryIQ in this scenario.

Question 1): “Are there any records for 253-899-6732?”

Question 2): “What are the top 10 most frequently called numbers for this person?”

Question 3): “Plot the known entities associated with these numbers on a map.”

Question 1) could be translated into a relatively simple query to a database of suspects. However, Questions 2) and 3) necessitate the LLM’s comprehension of the schemas of multiple data sources, including the database of suspects, phone call logs, and registered addresses. The LLM needs to translate these questions into complex queries comprising multiple steps and joining multiple tables across databases. **Figure 1** below captures a snapshot of QueryIQ, showing the agent question (Question 1 in this case), the response and SQL query generated by the LLM, and the result returned by the query during the initial step in a representative workflow empowered by QueryIQ. Exhibits 1-4 in Appendix illustrate additional steps of the inquiry by the agent, revealing some interesting patterns of social network connections and geospatial clusters among the frequent callers connected to the suspect. A close look at the returned query shows that the LLM goes beyond a simple SQL query to just retrieve a potential match of suspect to the given phone number, and generates a complex query that joins multiple tables to produce a complete entity profile of the matched suspect.

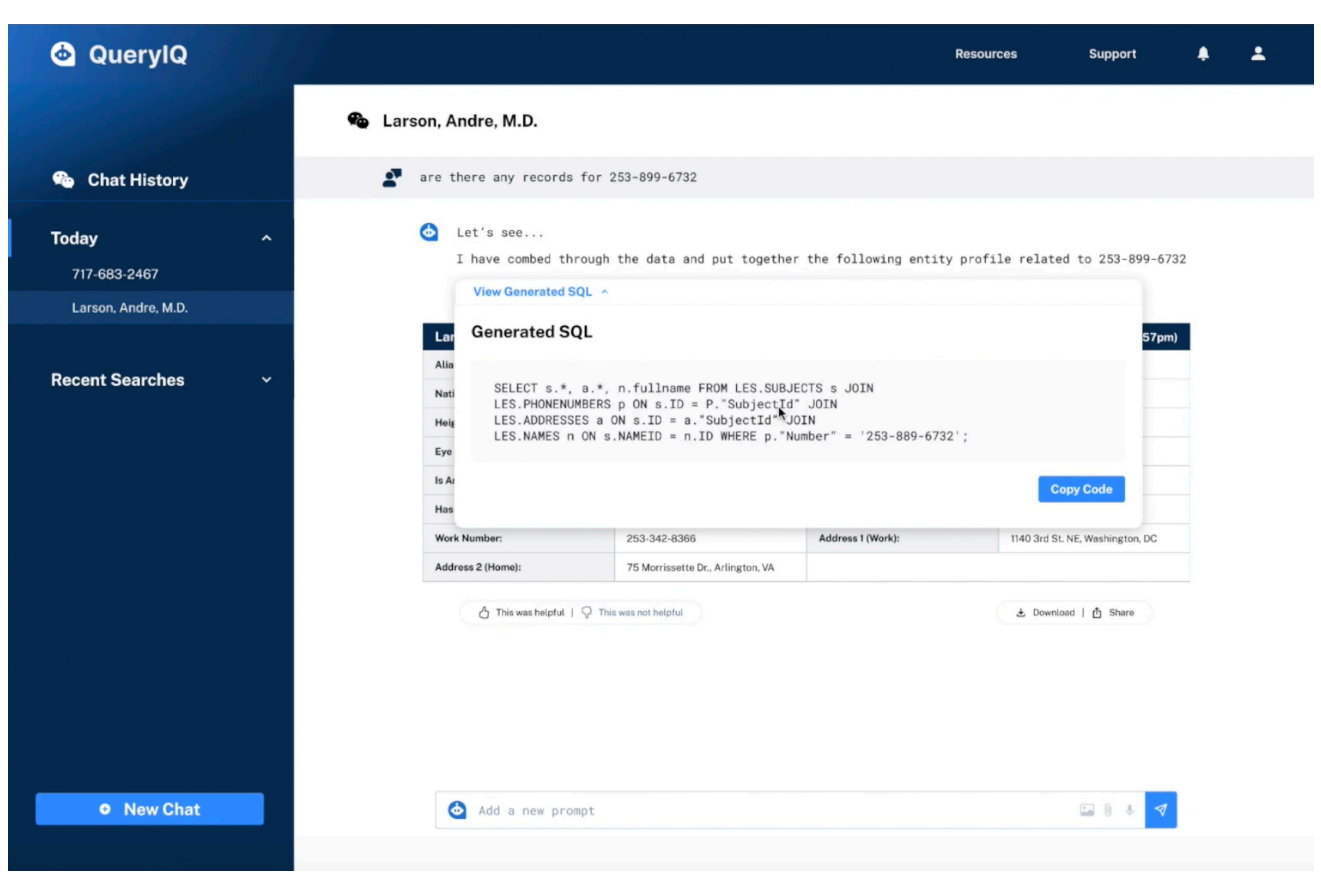

**Figure 1:** A snapshot of QueryIQ, illustrating text-to-query capability in a user workflow.

In contrast to data insight systems and user workflows empowered by text-to-query, users of conventional systems and workflows typically have to retrieve data and generates reports by triggering pre-defined and hardcoded queries and reporting functions, often through a click of a button or selection of filters on the user interface. It is impractical to anticipate and prepare for the myriad of complex questions (as those illustrated above) that can arise in dynamic real-world situations. Relying on standby data analysts with expertise in database queries and knowledge of data sources and schemas is clearly not scalable or efficient for addressing the multitude of complex questions posed by a large number of users at the speed of relevance.

# 3 Maturity Model

## 3.1 Overview

An overview of the proposed LLM maturity model focused on accuracy, consistency, and transparency is presented in **Table 1**, highlighting key characteristics in each category across increasing levels of maturity.

| Maturity | Level I | Level II | Level III | Level IV |
|---|---|---|---|---|
| Accuracy/ Efficacy | Perform basic translations of simple user questions into queries with limited functionality | Handle complex and broader user queries involving multiple databases and complex schemas or structures | Improved contextual understanding and handle nuanced user queries and implicit user intents effectively | Handle domain-specific queries and databases with a high degree of flexibility |
| Consistency/ Robustness | Significant variability and inconsistency in translating identical user questions | Occasional variability and inconsistency in translating identical user questions | Improved consistency and stability in translating equivalent/similar user questions | Highly consistent and stable in translating equivalent/similar user questions |
| Transparency/ Traceability | Minimum or no explanation or reasoning | Basic explanations provided with limited reasoning | Clear reasoning behind each step leading to the query | Advanced traceability with comprehensive logs for each decision |

**Table 1:** An overview of an LLM maturity model focused on accuracy, consistency, and transparency.

The Accuracy/Efficacy maturity category measures LLMs' capability in producing the correct queries across four maturity levels with general acceptability criteria as follows.

- Accuracy Threshold: Each maturity level has a defined accuracy threshold based on a chosen metric; for example, the average accuracy of LLMs in producing correct queries over a space of user questions.
- Handling of Complexity: Ability to handle complex queries, edge cases, linguistic nuances, and domain-specific terminologies and jargons.
- Test Dataset Performance: Evaluation based on a standardized test dataset representing a diverse range of user questions and database query scenarios.

The Consistency/Robustness maturity category measures LLMs' capability in producing consistent and stable results under variations of user questions, engineered prompts to LLMs, and underlying models across four maturity levels with general acceptability criteria as follows.

- Variation Tolerance Threshold: Each maturity level has a defined variation tolerance threshold based on a chosen metric; for example, the stability score of LLMs in producing syntactically and semantically correct queries under variations of inputs, such as LLM settings, user prompts, and linguistic variations in the question.

The Transparency/Traceability maturity category measures LLMs' capability in producing explanations, reasoning, and documentations of their decisions across four maturity levels with general acceptability criteria as follows.

- Interpretability: Availability and visibility of explanations and reasoning of decision making by the LLMs.
- Logging and Documentations: Comprehensiveness in various mechanisms of tracing, logging, and documenting data sources, queries, and metadata by exploring advanced user interface (UI) / user experience (UX) and adhering to government policies and regulations.

### 3.2 Detailed Presentation

#### 3.2.1 Accuracy/Efficacy Maturity

***Level I: Basic Text-to-Query LLMs***

Characteristics: At this stage, LLMs can generate basic queries from user questions, but exhibit limited accuracy and efficacy in dealing with complex and nuanced questions and joining information from multiple data sources and tables.

Example Query: "What are the names and ages of all employees in the Human Resources department?"

Acceptability Criteria as follows.

- Accuracy Threshold: Achieve a minimum of 60% (adjustable according to the domain and application) average accuracy in query generation.
- Handling Basic Queries: Successfully translate simple user questions to database queries.
- Test Dataset Performance: Reach the accuracy threshold on a standard benchmark dataset (e.g., Spider) with straightforward questions and simple database schemas or structures.

***Level II: Expanded Text-to-Query LLMs***

Characteristics: LLMs have an expanded capability to handle more diverse and complex queries with improved accuracy and efficacy, but exhibit limited capability in dealing with question context, nuances, and domain-specific jargons.

Example Query: "What are the top 10 products that have been ordered by customers located in the West region between January-June 2023, along with their respective order quantities?"

Acceptability Criteria as follows.

- Accuracy Threshold: Achieve a minimum of 80% (adjustable according to the domain and application) average accuracy in query generation.
- Handling Diverse and Complex Queries: Successfully translate complex user questions to database queries involving multiple joins, aggregations, nested structures, and other complex operations.
- Test Dataset Performance: Reach the accuracy threshold on a standard benchmark dataset with diverse questions and complex database schemas or structures.

***Level III: Contextually Aware Text-to-Query LLMs***

Characteristics: LLMs exhibit improved contextual awareness about prior queries, handling nuanced user questions and intent effectively, but are not tailored for specific domains.

Example Query: "List 5 employees who have shown best sales improvements."

It is worth noting that this is an example user query that implicitly conveys the user intent of comparing employee performances between two quarters based on sales figures, where the context could have been provided in a previous query "Who has the best sales figures in the last quarter?"

Acceptability Criteria as follows.

- Accuracy Threshold: Achieve a minimum of 90% (adjustable according to the domain and application) average accuracy in query generation.
- Contextual Understanding: Demonstrate the ability to accurately infer intent and context from user queries.
- Test Dataset Performance: Reach the accuracy threshold on customized evaluation datasets with various edge cases, sequences of example queries, and implicit user intents.

***Level IV: Domain-Specific Text-to-Query LLMs***

Characteristics: LLMs are highly specialized and tailored for domain-specific databases, exhibiting exceptional accuracy and efficacy, as well as expert-level understanding of domain-specific terminologies and databases.

Example Query: "Retrieve the recent maneuvers and reported positions of Red Force units within the training area Foxtrot-6 during the Exercise Thunderbolt, highlighting any detected incursions or potential threat patterns." Note that this is an example user query filled with terms and jargons widely used in military operations.

Acceptability Criteria as follows.

- Accuracy Threshold: Achieve above 90% (adjustable according to the domain and application) average accuracy in query generation.
- Domain Expertise: Demonstrate deep knowledge and proficiency in handling highly specialized queries related to specialized domain databases and understanding domain specific terminologies and jargons.
- Test Dataset Performance: Reach the accuracy threshold on domain-specific evaluation datasets with special terms and jargons.

### 3.2.2 Consistency/Robustness Capability

***Level I: Basic***

Characteristics: At this stage, LLMs exhibit minimum consistency in translating identical user questions under identical prompts and settings of LLMs, but significant variability and inconsistency under variations of prompts and settings.

Acceptability Criteria as follows.

- Variation Tolerance Threshold: Achieve a minimum of 80% stability in query generation for identical user questions under *identical* prompts and settings of LLMs.

***Level II: Enhanced***

Characteristics: LLMs exhibit enhanced consistency in translating identical user questions under variations of prompts and settings, but significant variability and inconsistency under linguistic variations of equivalent/similar user questions.

Acceptability Criteria as follows.

- Variation Tolerance Threshold: Achieve a minimum of 80% stability in query generation for identical user questions under *variations* of prompts and settings of LLMs.

***Level III: Advanced***

Characteristics: LLMs exhibit improved consistency and stability in translating equivalent/similar user questions.

Acceptability Criteria as follows.

- Variation Tolerance Threshold: Achieve a minimum of 60% stability in query generation for user questions under linguistic variations of equivalent or similar questions.

***Level IV: Expert***

Characteristics: LLMs are highly specialized and tailored for domain-specific databases, achieving exceptional accuracy and efficacy, as well as expert-level understanding of domain-specific terminologies and databases.

Acceptability Criteria as follows.

- Variation Tolerance Threshold: Achieve a minimum of 90% stability in query generation for user questions under linguistic variations of equivalent or similar questions.

### 3.2.3 Transparency/Traceability Capability

***Level I: Basic***

Characteristics: At this stage, LLMs exhibit basic transparency and traceability features such as basic logging of queries and responses, limited or no visibility into model decisions, and basic documentation of model architecture.

Acceptability Criteria as follows.

- Query Logging: Implement basic logging functionality to record queries and model responses.
- Basic Model Documentation: Provide a basic overview of the model architecture and decision-making processes.
- Compliance with Minimal Traceability Standards: Meet basic traceability standards set by regulatory bodies.

***Level II: Enhanced***

Characteristics: LLMs possess expanded transparency and traceability features such as enhanced query logging with metadata, intermediate-level interpretability, and clear documentation of data sources and preprocessing steps, offering more comprehensive insights into model decisions.

Acceptability Criteria as follows.

- Enhanced Query Logging: Include metadata (e.g., timestamps, user IDs, session information, etc.) in query logs for better traceability.
- Interpretability and Traceability Tools: Provide tools to gain insights into model decisions and confidence at an intermediate level.
- Detailed Data Documentation: Present comprehensive documentation detailing data sources used for training, preprocessing steps, and model training processes and hyperparameters.

***Level III: Advanced***

Characteristics: LLMs offer advanced transparency and traceability, providing clear reasoning behind each step leading to the query.

Acceptability Criteria as follows.

- Advanced Interpretability and Traceability Tools: Implement advanced techniques to provide detailed insights into model decisions and confidence, and provide observability/feedback UIs to facilitate users in staying on top of the loop and providing feedback.
- Comprehensive Documentation: Offer additional documentation covering model performance, biases, limitations, and areas of improvement beyond the previous levels.
- Transparent Disclosure Standards: Meet transparent disclosure standards set by regulatory bodies and ethical guidelines.

***Level IV: Expert***

Characteristics: LLMs showcase expert-level transparency and traceability, providing extensive insights and complete accountability with comprehensive logs for each decision.

Acceptability Criteria as follows.

- Complete Accountability Tools: Implement expert-level tools ensuring complete understanding and traceability of model decisions.
- Thorough Ethical Documentation: Offer in-depth documentation regarding societal impacts, ethical considerations, and potential consequences of model use.
- Robust Bias Mitigation Framework: Demonstrate robust mechanisms to identify, address, and mitigate biases, ensuring fairness and equity in model operations.

## 4 Future Work

To the best of authors' knowledge, this paper is the first to introduce an LLM maturity model for text-to-query applications. Focusing on the reliability and transparency aspects of LLMs, this paper contributes to a nuanced and comprehensive understanding of LLMs in this context. It lays the groundwork for the development and adoption of a comprehensive LLM maturity model, facilitating the assessment and deployment of LLMs in real-world applications.

An important future direction involves conducting extensive empirical studies and assessments on the top-ranked LLMs from various accuracy-dominated leaderboards. These assessments would compare the LLMs against the proposed maturity model, with a specific focus on identifying any significant gaps in terms of consistency and transparency in LLM performance, while also further opening the door to new research directions. Another natural extension of this work is to adapt the proposed maturity model to other applications and broaden its scope to encompass additional dimensions within the landscape of trustworthy AI.

## Acknowledgments

This work is initially presented at the Advantage DoD 2024: Defense Data & AI Symposium, https://www.ncsi.com/event/cdao/ hosted by the Chief Digital and Artificial Intelligence Office (CDAO) in the Department of Defense, February 20-22, 2024. The views and opinions expressed in this paper are solely those of the authors and do not necessarily reflect the official position of Expression or any government entity.

## Appendix

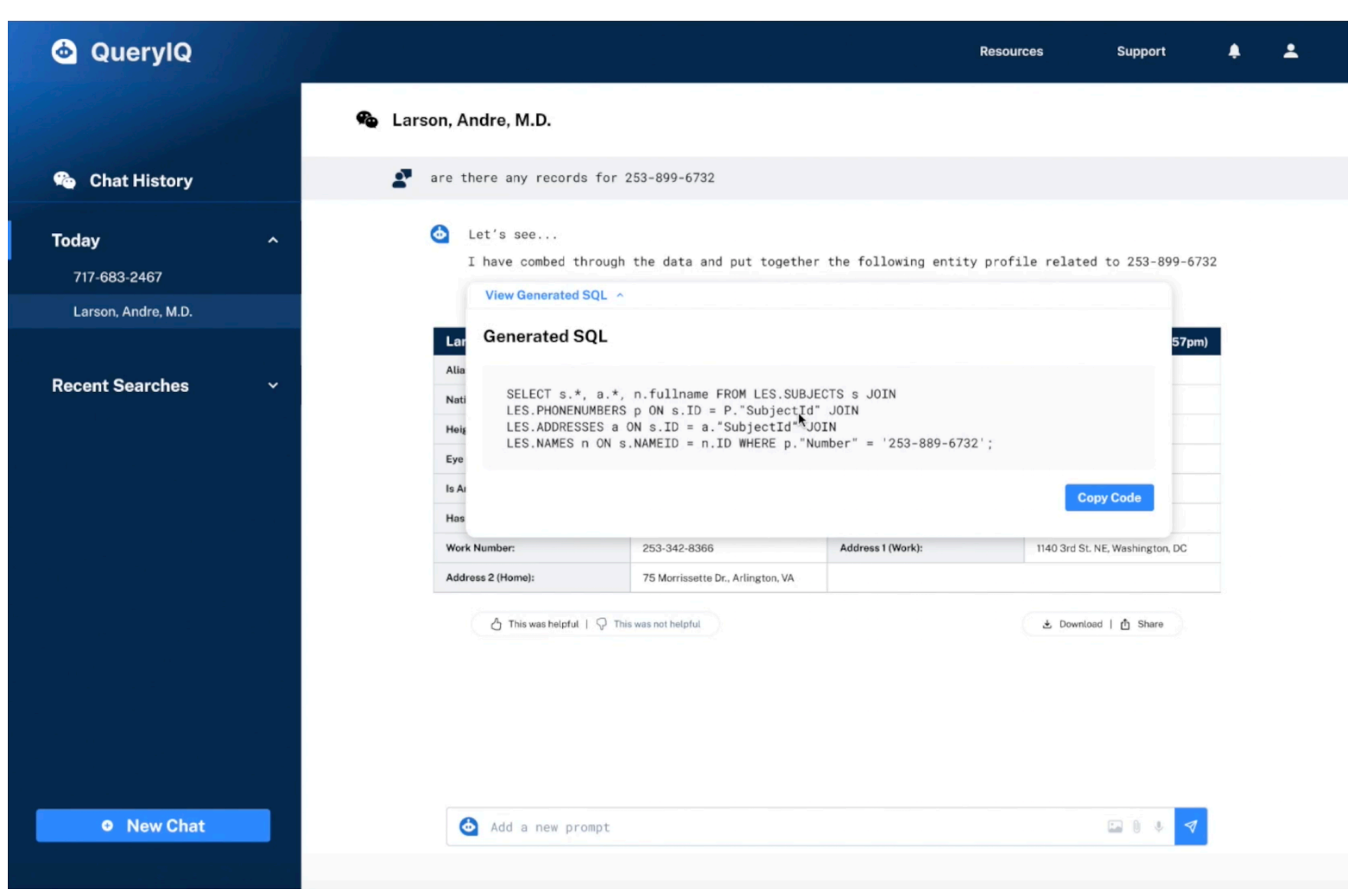

**Exhibit 1:** The LLM generates a complex query and produces an entity profile of the suspect linked to the given phone number during the initial step in a representative workflow empowered by QueryIQ.

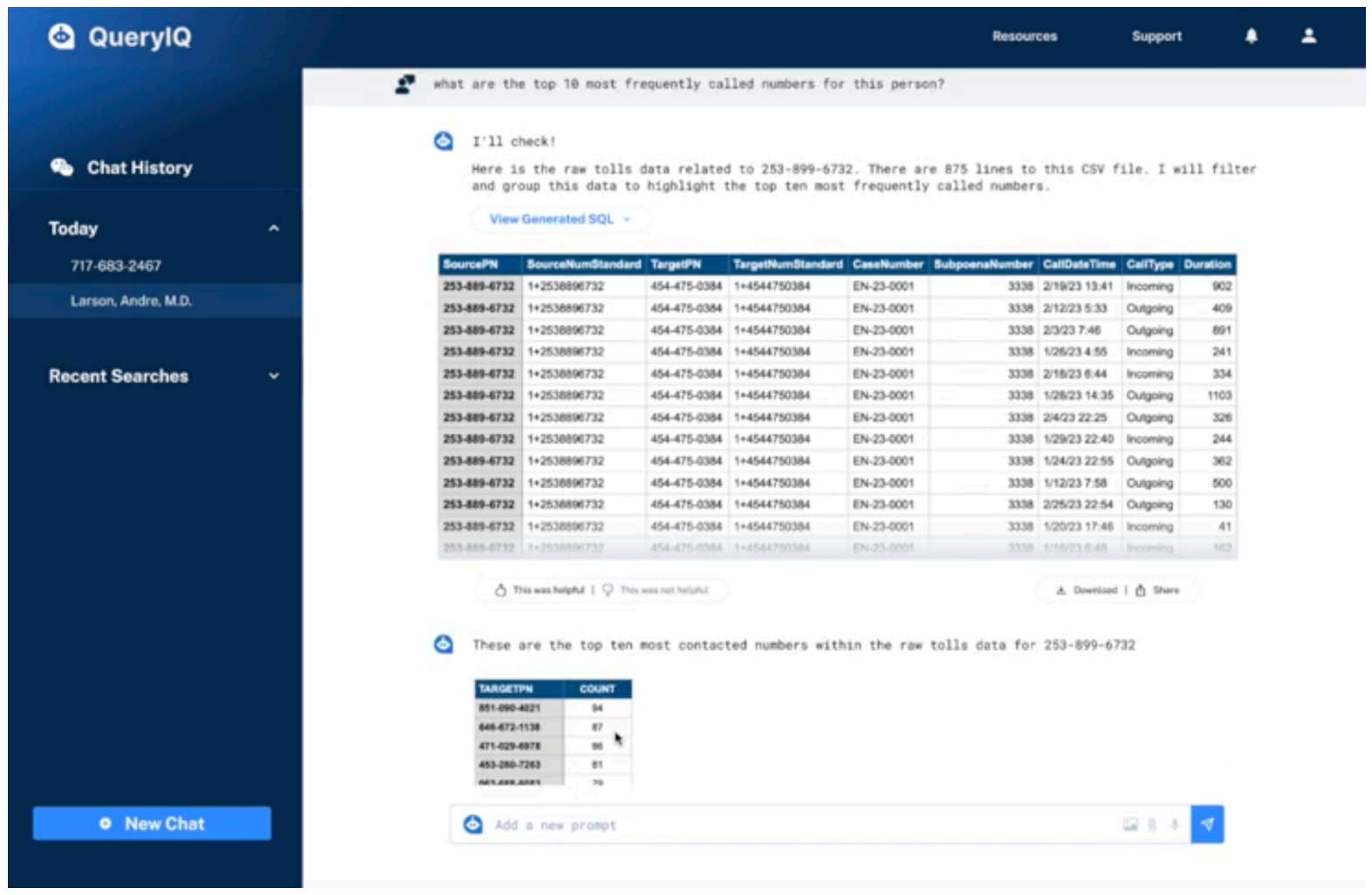

**Exhibit 2:** The LLM retrieves intermediate result of all tolls data related to the given phone number before producing the top 10 most contacted numbers for the suspect.

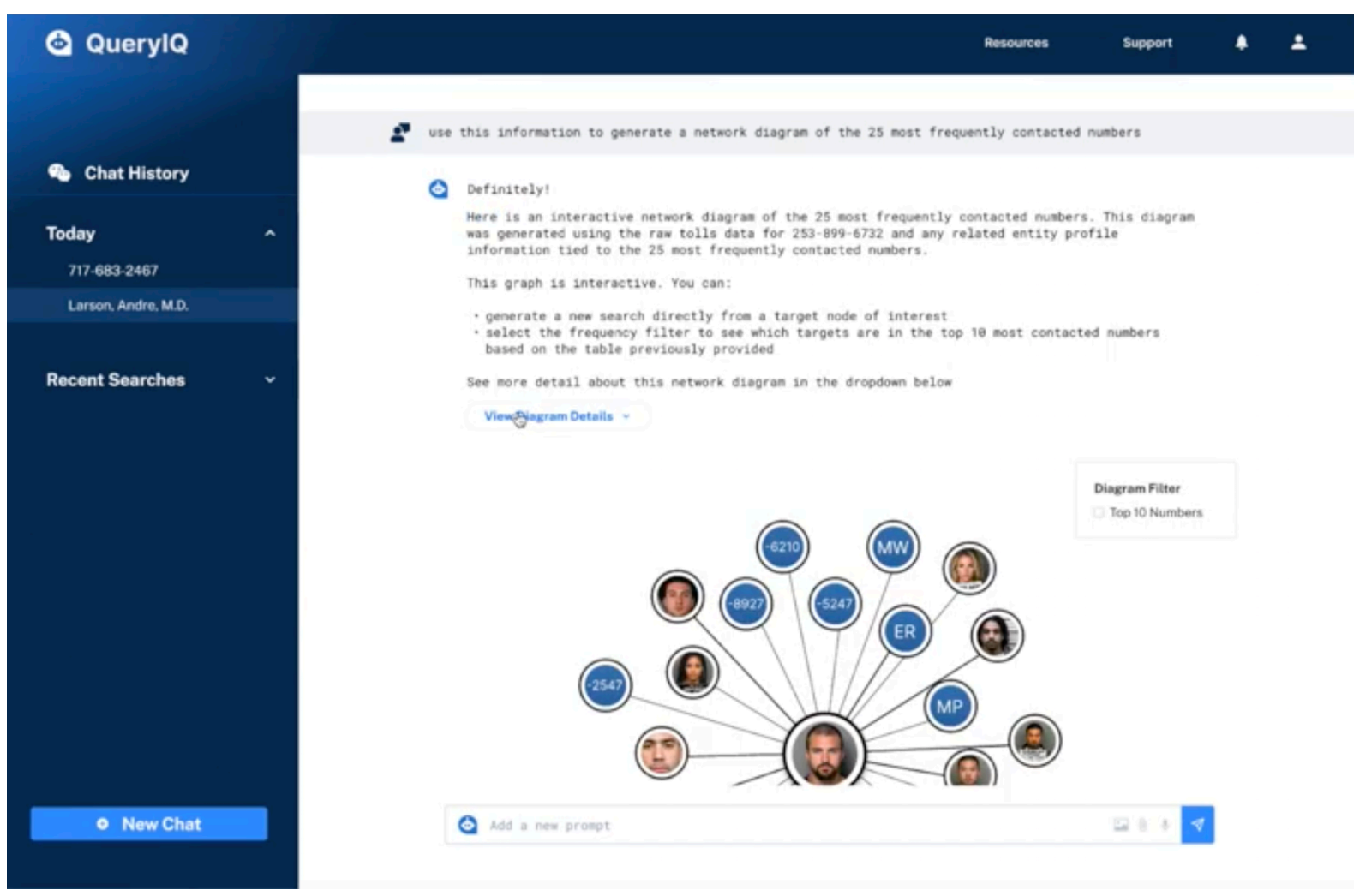

**Exhibit 3:** The LLM produces a network diagram based on the top 25 most contacted numbers.

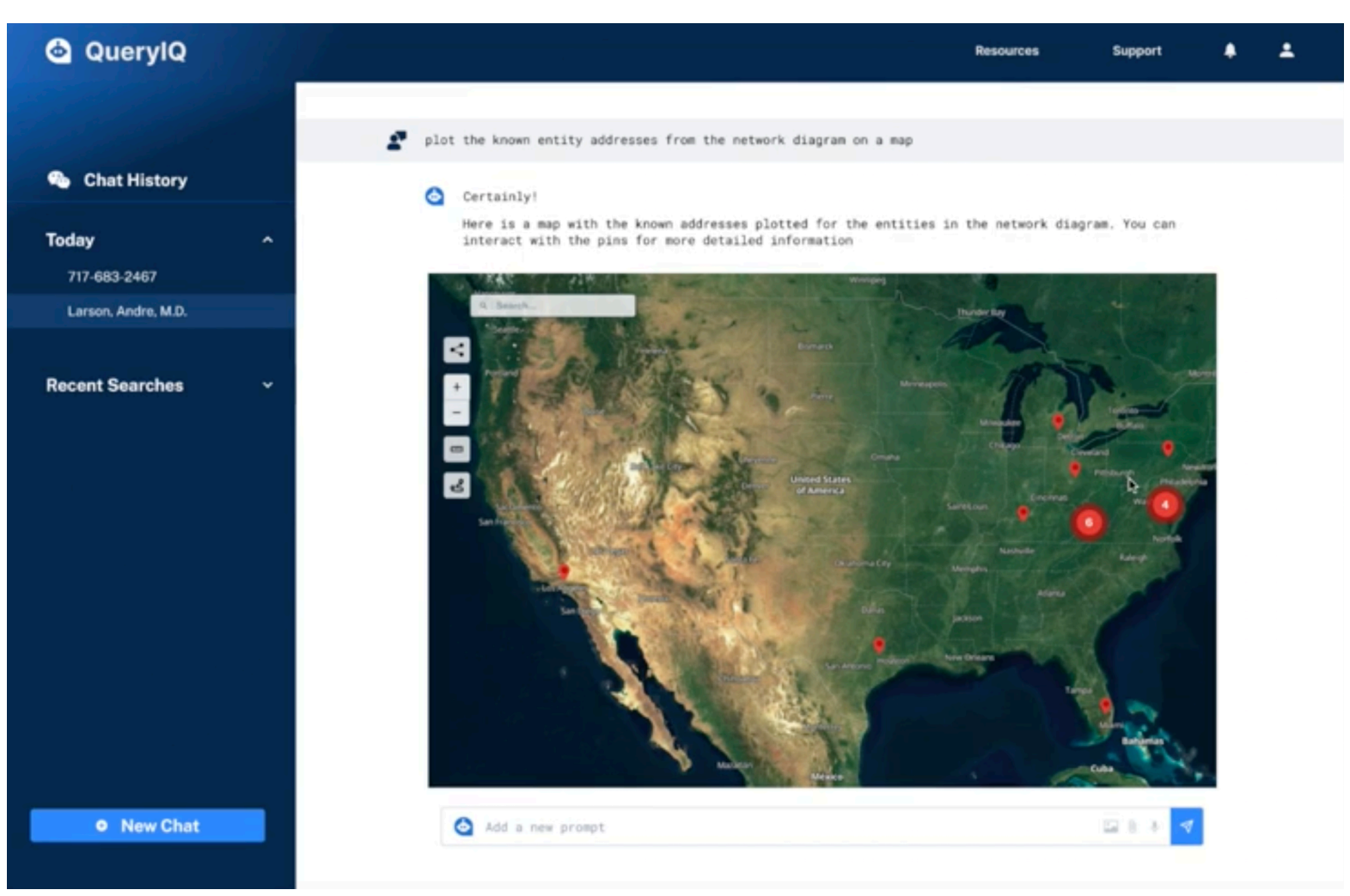

**Exhibit 4:** The LLM retrieves the addresses of the known entities from the network diagram and plots them on the map, revealing two clusters of entities in a region.